\documentclass[journal]{IEEEtran}
\usepackage{cite}
\usepackage{amsmath,amssymb,amsfonts}
\newtheorem{myDef}{Definition}
\usepackage{graphicx}
\usepackage{textcomp}
\usepackage{xcolor}
\usepackage{algorithm}
\usepackage{algpseudocode}
\usepackage{threeparttable}

\ifCLASSINFOpdf
\else
\fi
\begin{document}
%
\title{Deep Attributed Network Representation Learning via Attribute Enhanced Neighborhood}
%
%
%


\author{\IEEEauthorblockN{Cong Li, Min Shi, Bo Qu, Xiang Li}
	
	\thanks{This work is supported by National Natural Science Foundation of China (Grant No.71731004, No.62002184), and the Peng Cheng Laboratory Project of Guangdong Province PCL2018KP004. (\textit{Corresponding author: Xiang Li.})}
	\thanks{Cong Li, Min Shi and Xiang Li are with the Adaptive Networks and Control Lab, Department
		of Electronic Engineering, and with the Research Center of Smart Networks and Systems, School of Information Science and Technology, Fudan University, Shanghai 200433, China
		(e-mail: cong\_li@fudan.edu.cn; 18210720042@fudan.edu.cn; lix@fudan.edu).%
	}
	\thanks{Bo Qu is with Peng Cheng Laboratory, Shenzhen 518000, China
		(e-mail: qub@pcl.ac.cn).
	}
}

%
%

\markboth{Journal of \LaTeX\ Class Files}%
{Shell \MakeLowercase{\textit{et al.}}: Bare Demo of IEEEtran.cls for IEEE Journals}
%



\maketitle

\begin{abstract}
Attributed network representation learning aims at learning node embeddings by integrating  network structure and attribute information. 
It is a challenge to fully capture the microscopic structure and the attribute semantics simultaneously, where the microscopic structure includes the one-step, two-step and multi-step relations, indicating the first-order, second-order and high-order proximity of nodes, respectively. In this paper, we propose a deep attributed network representation learning via attribute enhanced neighborhood (DANRL-ANE) model to improve the robustness and effectiveness of node representations. The DANRL-ANE model adopts the idea of the autoencoder, and expands the decoder component to three branches 
to capture different order proximity. We linearly combine the adjacency matrix with the attribute similarity matrix as the input of our model, where the attribute similarity matrix is calculated by the cosine similarity between the attributes based on the social homophily. In this way, we preserve the second-order proximity to enhance the robustness of DANRL-ANE model on sparse networks, and deal with the topological and attribute information simultaneously. Moreover, the sigmoid cross-entropy loss function is extended to capture the neighborhood character, so that the first-order proximity is better preserved. We compare our model with the state-of-the-art models on five real-world datasets and two network analysis tasks, \textit{i.e.}, link prediction and node classification. The DANRL-ANE model performs well on various networks, even on sparse networks or networks with isolated nodes given the attribute information is sufficient.
\end{abstract}


%
\IEEEpeerreviewmaketitle

\section{Introduction}
%
%
%
%
\IEEEPARstart {N}{etworks} are generally utilized to explore and model complex systems, such as online social networks and citation networks, where an entity is represented as a node and the interaction between two entities is represented as an edge. Network analysis is an effective way to gain insight into different aspects of complex systems, which derives many machine learning applications, such as the online advertisement targeting \cite{Wang2017} and anomaly detection \cite{Chandola2009}. Hence, identifying effective features of nodes (or edges) in a network is essential. However, traditional methods tend to manually mine the specific domain features depending on the expert experience, which not only require the high cost of labor and time, but also limit the scalability of models on different prediction tasks \cite{Grover2016}. Network representation learning (NRL) \cite{Perozzi2014} as an alternative of automatic feature mining has been proved to be beneficial to various network analysis tasks, such as the node classification \cite{Sen2008}\cite{Kazienko2012}, link prediction \cite{Liben-Nowell2007}, clustering \cite{Narayanan2007} and visualization\cite{Maaten2008}.

Early NRL methods are mostly based on the matrix factorization\cite{Yan2007}. 
To reduce the computational complexity on large-scale networks, inspired by word modeling, Perozzi \textit{et al.} propose DeepWalk\cite{Perozzi2014}.
Notably, the above-mentioned shallow model design 
cannot well capture the highly non-linearity that is universal in the networks, 
which would lead to the sub-optimal network representation \cite{Wang2016}. 
Then, many deep models, such as structural deep network embedding (SDNE)\cite{Wang2016}, have emerged.

Considering only the network structure is not enough to learn the informative and accurate node representations, especially when the network structure is sparse. 
Social science theories like the homophily \cite{Marsden1993} \cite{McPherson2001} and social influence theory \cite{Marsden1988} suggest that there is a strong correlation between the structure and the attributes. 
Therefore, many studies focus on attributed NRL\cite{Gao2018} \cite{Liao2018} \cite{Zheng2019}, which mainly learn the consistent node representations from the network structure and node attributes. Nevertheless, the above models are likely susceptible to the sparsity of one of the
heterogeneous information sources. Afterwards, the deep coupling paradigm is introduced to enhance the robustness of the 
representations. Attributed network representation learning (ANRL)\cite{Zhang2018} is one of the representative examples. However, the ANRL model could be affected by the characteristics of the local network structure. In addition, the previous attributed NRL methods rarely preserve all the microscopic structural information, \textit{i.e.}, the first-order, second-order, and high-order proximity\cite{Zhang2018N}, together, where these proximities indicate the one-step, two-step, and multi-step relationship between two nodes. Nevertheless, explicitly taking full advantage of the microscopic 
structure tends to be essential for learning the network representation \cite{Cao2015}.

To utilize the microscopic structural information as well as the node attributes, we propose a novel deep coupling attributed NRL model, namely, the deep attributed network representation learning via attribute enhanced neighborhood (DANRL-ANE) model. The proposed model consists of three coupled modules, \textit{i.e.}, the self-built first-order proximity preserved, attribute enhanced neighborhood autoencoder and community-aware skip-gram module, which preserve the first-order, second-order, and high-order proximity, respectively. The three modules share connections to the encoder. Especially, we model the attributes based on the social homophily, and incorporate the attribute semantics into the adjacency matrix to enhance the direct neighborhood of each node.

To summarize, our main contributions are as follows:

(i) We propose a deep three-part coupling model, DANRL-ANE, which learns the robust and effective node representations by jointly mining the microscopic structure and node attributes. The attributes are preprocessed to be used as the input of our model together with the adjacency matrix, which is advantageous to obtain the accurate second-order proximity.

(ii) We construct the self-built first-order proximity preserved module, which innovatively extends the sigmoid cross-entropy loss function for capturing the local pairwise relationship between node pairs on undirected and unweighted networks.

(iii) Our proposed model is not only good for the machine learning tasks that benefit from the pairwise properties between nodes, \textit{i.e.}, the link prediction and node classification, but also not susceptible to the sparsity and neighborhood distribution of either the structure or the attributes. Moreover, our model even can deal with the networks with isolated nodes when we obtain the sufficient node attributes.

The rest of the paper is organized as follows. We discuss the related work in Section II, and introduce the preliminaries involved in the paper in Section III. We give the detailed description of our model in Section IV, and then show the experimental settings and results in Section V. Finally, we conclude the paper in Section VI.


\section{Related Work}
The network representation is first proposed as the part of dimensionality reduction technologies \cite{Goyal2018}\cite{Roweis2000}\cite{Tenenbaum2000} in the early 20th century. However, the early methods suffers from both computational and statistical performance drawbacks \cite{Grover2016}. Afterwards, Perozzi \textit{et al.}\cite{Perozzi2014} generalize the advancements in language modeling to large-scale networks. A large number of related excellent algorithms have been proposed. For example, DeepWalk\cite{Perozzi2014} uses the uniform sampling to collect the node sequences\cite{Cao2015}. 
Node2vec \cite{Grover2016} extends the DeepWalk model, which 
captures the diversity of connectivity patterns 
in a network. 
Intuitively, the skip-gram-based methods 
capture the high-order proximity\cite{Nguyen2018}. LINE\cite{Tang2015} 
utilizes different carefully designed objective functions to preserve the first-order and second-order proximity. However, all of the above methods cannot preserve the different $k(k\geq3)$-step relationships in distinct subspaces\cite{Cao2015}. Therefore, Cao \textit{et al.} propose GraRep\cite{Cao2015}, which concatenates all the local $k(k\geq3)$-step representations 
as the representations of nodes. The mentioned methods all utilize the network structure only to learn network representation. Besides the structure, nodes in the real world are usually affiliated with various attributes.

Then, researchers begin to focus on mining the network features from attributed networks, such as GAT2VEC \cite{Sheikh2019} and SANE \cite{Liu2019}. 
To further capture the highly non-linearity, some algorithms, such as DANE\cite{Gao2018}, ASNE\cite{Liao2018} and MDNE\cite{Zheng2019}, have been recently designed based on the deep learning technologies, which all model the network structure, encode the attribute information, and then depend on the strong correlation between the structure and the attributes to obtain the consistent network embedding. Giving an example, DANE\cite{Gao2018} utilizes the autoencoder to preserve the high-order structural proximity 
and attribute semantics,  
the joint probability to capture the first-order proximity from the structure and attributes, and the likelihood estimation to learn the node embeddings by jointly the structure and attributes. 
The above methods might be susceptible to the sparsity of either the structure or the attributes. For learning the robust representations, 
Zhang \textit{et al} \cite{Zhang2018} propose a deep coupling model ANRL, which preserves the second-order and high-order proximity from the topological structure. On the basis of the encoder part, ANRL constructs a neighbor enhancement autoencoder module, and designs an attribute-aware skip-gram module. Nevertheless, the design of the autoencoder makes ANRL limited by the choice of datasets.

In summary, attributed NRL is still an open problem in at least two aspects as follows: (1) the network structure and node attributes are two heterogeneous information sources, we need to consider how to preserve their characteristics in a vector space; (2) the first-order, second-order and high-order proximity define different neighborhood relations among directly or indirectly connected nodes. To capture local closeness proximities could help preserve the entire microscopic structure features of original networks, yet how to design a proper model is a challenge. Here, we propose the DANRL-ANE model under the paradigm of deep coupling, in which three coupled modules are designed to capture the different order proximity. Especially, the attribute information is mined as the supplement of the adjacency matrix.

\section{Preliminaries}
In this section, we first briefly introduce two types of node attributes, notations and definitions which are used in this work. Then, the schematic of attributed network representation learning is given.

\subsection{Node Attributes}
The node attributes refer to the auxiliary information used to describe a node besides the network structure. For instance, in social networks, personal information such as age, gender and hobbies can be used as attributes. As declared in \cite{Liao2018}, regardless of the semantics, the attributes could be categorized into two types: the discrete attributes and continuous attributes.

$\bullet$ \textit{Discrete attributes.} The typical example of the discrete attributes is the categorical attributes, which can be transformed into the binary vectors via one-hot encoding.

$\bullet$ \textit{Continuous attributes.} The continuous attributes naturally exist in social networks. They could be artificially generated from the transformation of the categorical variables. The continuous attributes can be represented as the real-valued vectors after being preprocessed. For example, in the document modeling, after obtaining the bag-of-words representation of a document, it is common to transform it to a real-valued vector via TF-IDF to reduce the noises \cite{Liao2018}.

Our proposed model DANRL-ANE is suitable for the networks with either discrete attributes or continuous attributes.

\subsection{Notations}
Let $G=(V,E,A,X)$ be an attributed information network, where $V=\{v_i, ..., v_n\}$
denotes a set of nodes, $E \subset (V \times V) $ denotes a set of edges among nodes, $A$ is the adjacency matrix and $X$ is the node attribute matrix. In the adjacency matrix $A$, if there is an edge between nodes $v_i$ and $v_j$, $a_{ij} > 0 $, particularly, if the network is unweighted, $a_{ij} = 1 $; otherwise, $a_{ij} = 0$. If the network is undirected, $a_{ij}=a_{ji}$. In the node attribute matrix $X$, the element $x_{ik}$ indicates the value of node $v_i$ on the attribute $k$. In this work, we focus on the undirected and unweighted networks.

\subsection{The Closeness Proximity}
We here introduce the definition of the first-order, second-order and high-order proximity involved in our model.

\begin{myDef}
	\textbf{\textit{First-order proximity}}
\end{myDef}

The first-order proximity describes the pairwise proximity between nodes\cite{Wang2016}. For each node pair $(v_i,v_j)$, if there is an edge between them, the first-order proximity between nodes $v_i$ and  $v_j$ is $a_{ij}$; otherwise, the first-order proximity between nodes $v_i$ and $v_j$ is $0$.

\begin{myDef}
	\textbf{\textit{Second-order proximity}}
\end{myDef}

The second-order proximity between a pair of nodes describes the proximity of the neighborhood structure of the node pair\cite{Wang2016}. Let $A_{i}=[a_{i1}, a_{i2},...,a_{in}]$ denote the first-order proximity between node $v_i$ and all other nodes, then the second-order proximity between nodes $v_i$ and $v_j$ is decided by the similarity measure, such as cosine similarity, between $A_{i}$ and $A_{j}$. Notice that the second-order proximity captures the $2$-step relation between node pairs, which could be measured by the $2$-step transition probability from node $v_i$ to node $v_j$, equivalently\cite{Zhang2018N}.

\begin{myDef}
	\textbf{\textit{High-order proximity}}
\end{myDef}

Compared with the second-order proximity, the high-order proximity captures the more global structure, which explores the $k$-step $(k \ge 3)$ relation between node pairs\cite{Zhang2018N}. The high-order proximity could be measured by the $k$-step$(k \ge 3)$ transition probability from node $v_i$ to node $v_j$.

\begin{figure}[h]
	\setlength{\abovecaptionskip}{0.cm}
	\setlength{\belowcaptionskip}{-0.cm}
	\centering
	\includegraphics[width=3.5in]{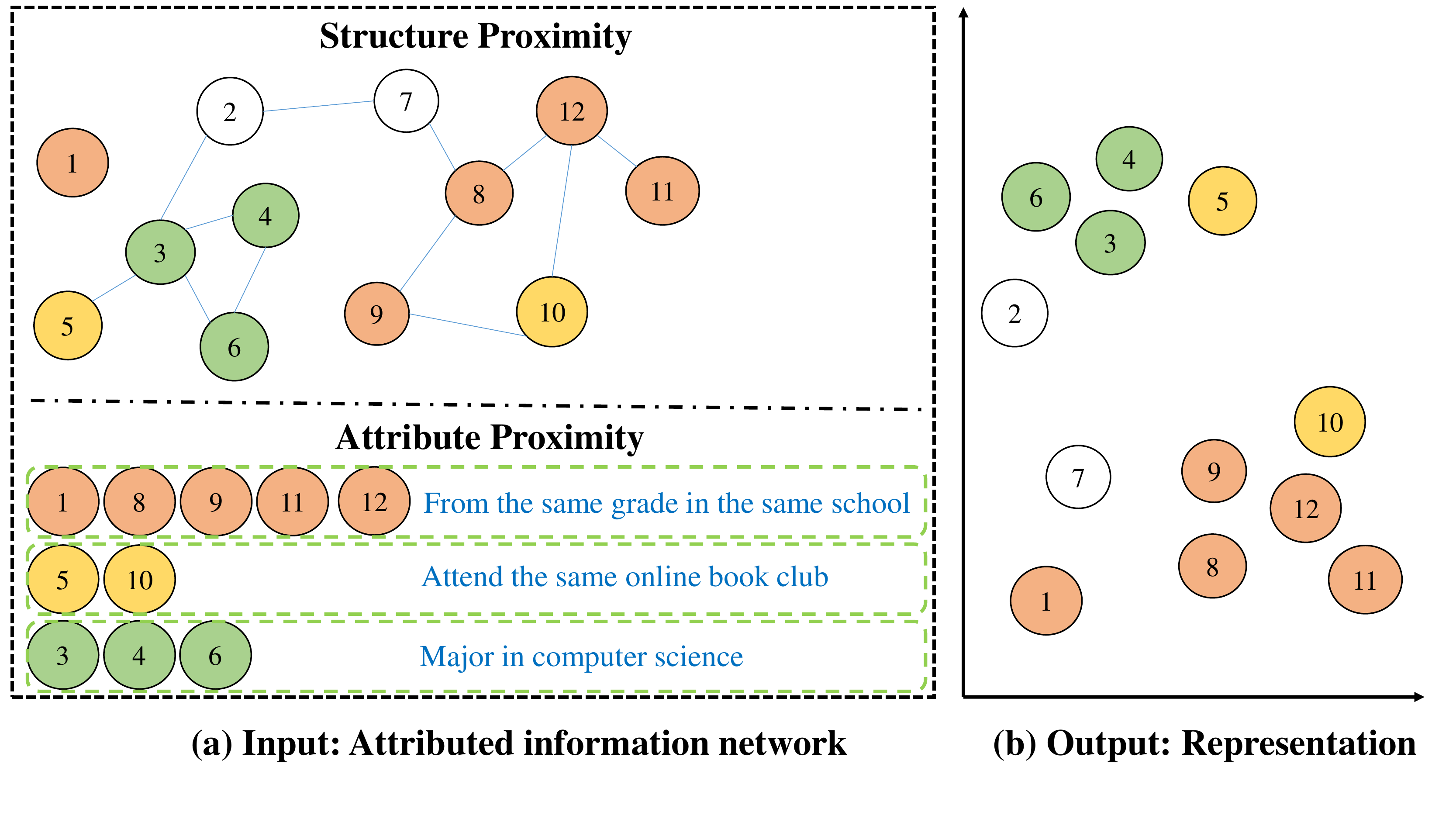}
	\caption{An illustration of attributed network representation learning. A social network with node attribute information is the input, and the vector representations of nodes is the output. \textit{Input}: the numbered nodes denote the users, the edges between nodes represent the social relations between users, and the same color nodes represent that the users have the similar attribute information. \textit{Output}: the vectors preserve the network structural information and attribute semantics.}
	\label{description}
\end{figure}

\subsection{Attributed network representation learning}
The goal of the attributed network representation learning is that 
with a given attributed information network $G=(V,E,A,X)$, learning a mapping function makes the whole network embedded into a new low-dimensional vector space, namely, $f:G\to Y\in \mathbb{R}^{n \times d}$, where $d$ denotes the dimension of embedding. Then, each node can be represented by a low-dimensional and dense 
vector. 
The vectors store the relationship information between each node and the other nodes, and record the attribute semantics of the nodes.  
Taking the node representations as the input is beneficial for the subsequent machine-learning-based network analysis tasks. 
A schematic of attributed network representation learning is shown in Fig. \ref{description}. It can be seen that the nodes close to each other in the original network and/or nodes with the similar attributes are also close to each other in the new vector space.

\section{The DANRL-ANE Model}

\subsection{Overview}
\begin{figure}[h]
	\setlength{\abovecaptionskip}{0.cm}
	\setlength{\belowcaptionskip}{-0.cm}
	\centering
	\includegraphics[width=4.4in]{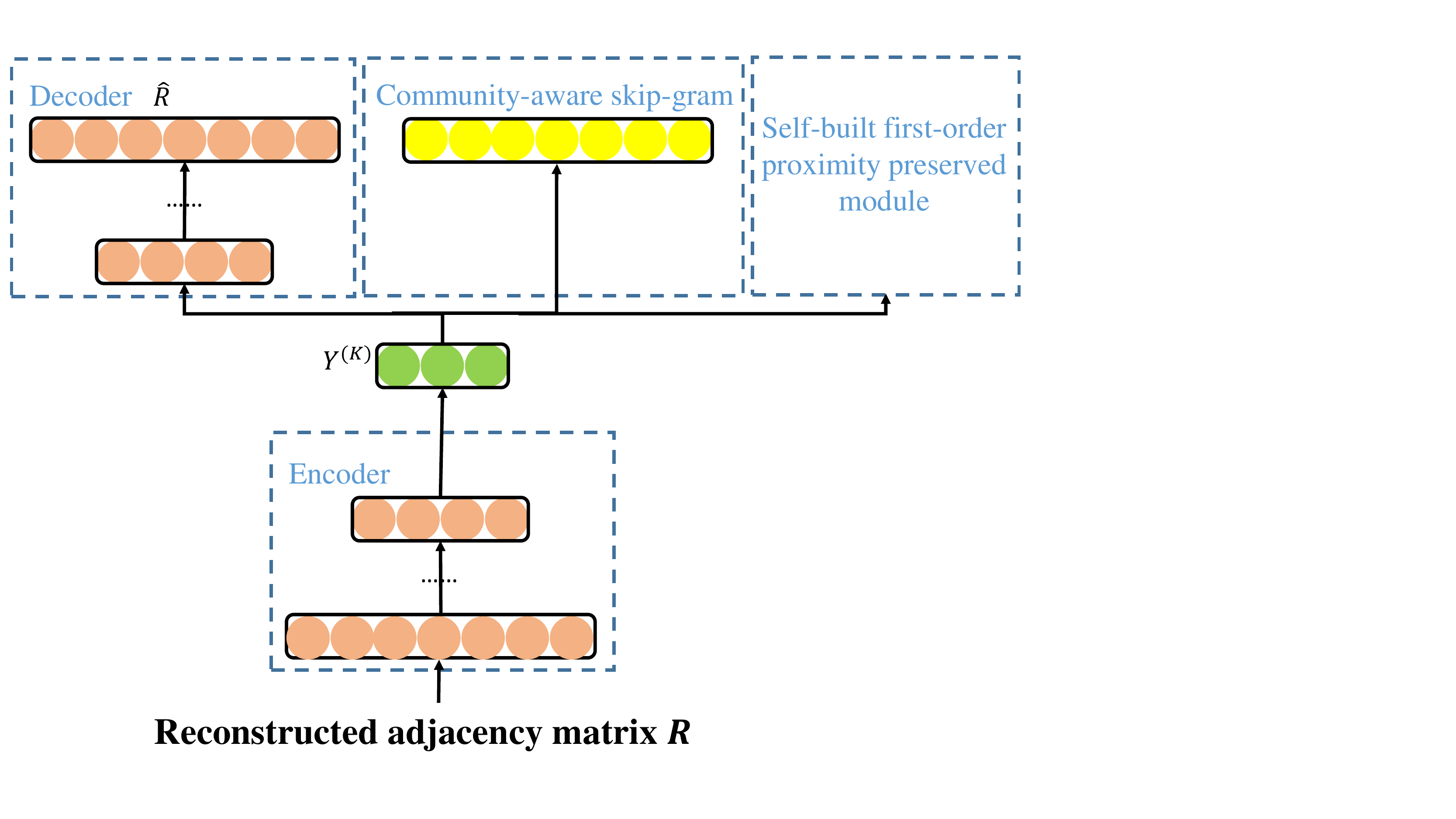}
	\caption{The architecture of DANRL-ANE model. DANRL-ANE is a three-part coupling model: the left is the attribute enbanced neighborhood autoencoder module, 
		the middle is the community-aware skip-gram module 
		and the right is the self-built first-order proximity preserved module. 
	}
	\label{model_2}
\end{figure}

The proposed DANRL-ANE model is a deep three-part coupling model, which consists of the self-built first-order proximity preserved module, the attribute enhanced neighborhood autoencoder module and the community-aware skip-gram module. Fig. \ref{model_2} shows the framework of DANRL-ANE model. 
The input of the encoder is the reconstructed adjacency matrix, which is obtained by integrating the node attributes and adjacency matrix. The self-built first-order proximity preserved module captures the direct relations between nodes, the attribute enhanced neighborhood autoencoder module reconstructs the target neighbors of nodes to learn the relations between the neighborhoods of two nodes, and the community-aware skip-gram module is trained on the linear node sequences to preserve the high-order relations. By training the three modules iteratively until the model converges, the final node representations are obtained, namely, the representation output of the autoencoder.  

\subsection{Preprocessing}
Considering that the attributes can provide direct evidence for the similarity measurement between nodes from the attribute level, we propose to construct an attribute similarity matrix. 
The formal description is given as follows.

\textbf{1) Attribute Similarity Matrix $X^{(S)} \in \mathbb{R}^{n \times n} $} 

Every row $X_{i}$ of an attribute matrix $X$ represents the attribute information of the corresponding node $v_i$. The attribute similarity $x_{ij}^{(S)}$ between nodes $v_i$ and $v_j$ could be calculated based on the similarity measurement methods. 
Inspired by the previous work \cite{Strehl2000}, we utilize the cosine similarity to calculate the attribute similarity 
\begin{equation}
x_{ij}^{(S)}=CosineSimilarity(X_{i},X_{j})=\frac{X_{i}X_{j}^{\top}}{|X_{i}||X_{j}|} .
\end{equation}

Furthermore, we intend to combine the adjacency matrix $A$ and the attribute similarity matrix $X^{(S)}$ into a new reconstructed adjacency matrix to strengthen the relationship between nodes.

\textbf{2) Reconstructed Adjacency Matrix $R \in \mathbb{R}^{n \times n} $}

Different from the attribute similarity matrix $X^{(S)}$, the adjacency matrix $A$ describes the similarity between nodes from the structure level. 
By setting the hyperparameters $\eta$ and $\psi$, we linearly combine the adjacency matrix $A$ and attribute similarity matrix $X^{(S)}$ to build the reconstructed adjacency matrix $R$ 
\begin{equation}
\label{E1}
R =\eta A+\psi X^{(S)} .
\end{equation}

\subsection{Model Design}
\textbf{1) A Joint Optimization Framework of the DANRL-ANE Model}

The entire model consists of three coupled modules, 
which share the same encoder component. The encoder aims at mapping the input data into the representation space by one or multiple layers of non-linear functions. In our model, the input-output relationship of each layer of the encoder 
is defined as
\begin{align}
\begin{split}
\label{E20}
y_{i}^{(1)} &=\delta (R_{i}W^{(1)} + b^{(1)})\\
y_{i}^{(k)} &=\delta (y_{i}^{(k-1)}W^{(k)} + b^{(k)}), k\in\{2,...,K\} ,
\end{split}
\end{align}
where $R_{i}$ is the $i$-th row of the input matrix $R$, and records the reconstructed neighbor relationship of node $v_i$. 
The symbol $\delta(.)$ denotes the non-linear activation functions, and we choose the best suitable one based on their performance on different tasks and datasets. 
The model parameters $W^{(k)}$ and $b^{(k)}$ represent the transformation matrix and bias vector in the $k$-th layer, respectively, and $K$ is the layer number of the encoder. 

The joint optimization objective of the DANRL-ANE model 

\begin{equation}
\label{E8}
L=L_{sg}^{NS} + \alpha L_{ae}^M + \beta L_{FoP} + \gamma L_{reg}
\end{equation}
with

\begin{equation}
\label{E9}
L_{reg}=\frac{1}{2}\sum_{k=1}^{K}(\Vert W^{(k)}\Vert_F^2 + \Vert \hat W^{(k)}\Vert_F^2) ,
\end{equation}
where $L_{FoP}$, $L_{ae}^M$, $L_{sg}^{NS}$ are the loss function of the  self-built first-order proximity preserved module, the attribute enhanced neighborhood autoencoder module and the community-aware skip-gram module, respectively, as well as $\alpha$ and $\beta$ are the hyperparameters that balance the effect of the modules. 
The hyperparameter $\gamma$ is the coefficient of the $l_2$ norm regularizer $L_{reg}$ that is used to prevent overfitting. 
The number of layers in the encoder and decoder is $K$. The matrix $W^{(k)}$ and $\hat W^{(k)}$ represents the weight matrix of the encoder and decoder in the $k$-th layer, respectively.

We optimize Equation (\ref{E8}) by the stochastic gradient algorithm adopted by ANRL \cite{Zhang2018}. 

Next, we give the detailed description of each module.

\textbf{2) Self-built First-order Proximity Preserved Module}
The first-order proximity can reveal the similarity between nodes intuitively and simply. We here propose a self-built first-order proximity preserved module. To model the first-order proximity, inspired by the LINE \cite{Tang2015} and DANE \cite{Gao2018} model, 
we define the joint probability $w_{ij}(v_i, v_j)$ between nodes $v_i$ and $v_j$ with the sigmoid function $\sigma(x)=\frac{1}{1+\exp (-x)}$. Let $x=y_{i}^{(K)}y_{j}^{(K)^ \mathrm{ T }}$, then we obtain 
\begin{equation}
\label{E5}
w_{ij}(v_i, v_j)=\frac{1}{1+\exp (-y_{i}^{(K)}y_{j}^{(K)^ \mathrm{ T }})} ,
\end{equation}
where $y_{i}^{(K)}$ and $y_{j}^{(K)}$ denote the representation of nodes $v_i$ and $v_j$, respectively.


Essentially, the first-order proximity on undirected and unweighted networks describes the existence or non-existence of edge between node pairs, which is equivalent to a binomial classification problem. The sigmoid cross-entropy loss function is a typical objective function for a binomial classification, 
which is defined as
\begin{equation}
\label{E6}
L_{ScE}=-[t\log p_s(s) + (1-t)\log (1-p_s(s)) ] ,
\end{equation}
where $t$ represents the label of a sample, and can be either $1$ or $0$. If the sample belongs to the positive class, $t=1$, else $t=0$. The probability $p_s(s)$ indicates the possibility that the sample is predicted to belong to the positive class, and is calculated by the sigmoid function. Here, $t=a_{ij}$ and $p_s(s)=w_{ij}(v_i, v_j)$, equivalently. 
Because only the information about the existing edges is useful for the network representation, we set $t=1$, and take the first term of Equation (\ref{E6}) into consideration. We average all the losses when the edges exist, and thus obtain the objective of the self-built first-order proximity preserved module
\begin{equation}
\label{E7}
L_{FoP}=\overline{\sum_{a_{i,j}=1}(-\log w_{ij}(v_i, v_j))} ,
\end{equation}
where $a_{ij}$ represents the element of the $i$-th row and the $j$-th column of the adjacency matrix $A$.

\textbf{3) Attribute Enhanced Neighborhood Autoencoder Module}

The deep autoencoder model is widely used to mine the proximity between the neighborhood structure of node pairs \cite{Wang2016}, since it could smoothly capture the data manifolds, and preserve the similarity between samples\cite{Ruslan2009}.  

The autoencoder consists of two parts, \textit{i.e.} the encoder and decoder. The decoder is the inverse calculation process of the encoder. 
The representation in the representation space is mapped into the reconstruction space in the decoder process.

The purpose of the autoencoder is to minimize the reconstruction error between the input data and the reconstructed data, so that the abstract representation of the mid-layer output can capture the manifold structure in the input data. To be specific, the objective of the autoencoder is 
\begin{equation}
L_{ae} =\sum_{i=1}^n||\hat{R_{i}}-R_{i}||_2^2 ,
\end{equation}
where $n$ is the number of the nodes in the networks, and $\hat{R_{i}}$ represents the  reconstructed output of the input data $R_{i}$. 

Although the direct neighborhood of each node gets enhanced after integrating the attribute semantics and adjacency matrix, the reconstructed adjacency matrix $R$ could still be a sparse matrix, that is, the number of non-zero elements is far less than that of zero elements. 
A sparse input matrix could further make the autoencoder tend to preserve more zero elements. Since the non-zero elements record the connections between nodes, it is more essential to preserve non-zero elements as much as possible instead of zero elements. 
Inspired by SDNE \cite{Wang2016}, we employ the Hadamard product as the penalty factor, and extend the loss function of the autoencoder. 
The modified objective is
\begin{equation}
\label{E3}
L_{ae}^M=\sum_{i=1}^n||(\hat{R_{i}}-R_{i})\odot b_{i}||_2^2 ,
\end{equation}
where $\odot$ indicates the Hadamard product, $b_{i}=\{b_{i,j}\}_{j=1}^n$. If $R_{i,j}=0$, $b_{i,j}=1$, else $b_{i,j}=\chi >1$.

The attribute enhanced neighborhood autoencoder module takes each row vector of the reconstructed adjacency matrix $R$ as the sample input, where the row vector denotes the neighbor structural information with the attribute semantics of the corresponding node. 
In other words, the autoencoder could preserve the second-order proximity. 

\textbf{4) Community-aware skip-gram module}

Inspired by DeepWalk \cite{Perozzi2014}, we design the community-aware skip-gram module for capturing the high-order proximity in this work. 

To reduce the time complexity, we adopt the node sequences sampling procedure performed by node2vec\cite{Grover2016}, where the return parameter $p_n=1.0$ and in-out parameter $q_n=1.0$, and use the negative sampling to approximate the following loss function
\begin{align}
\begin{split}
\label{E2}
L_{sg} &=-\sum_{i=1}^{n}\sum_{c\in C}\sum_{-b\le j\le b,j\ne 0}\log p(v_{i+j}|R_{i})\\
&=-\sum_{i=1}^{n}\sum_{c\in C}\sum_{-b\le j\le b,j\ne 0}\log \frac{\exp(h_{(i+j)}^{\prime }y_{i}^{(K)^ \mathrm{ T }})}{\sum_{f=1}^{n}\exp(h_{f}^{\prime }y_{i}^{(K)^ \mathrm{ T }})} ,
\end{split}
\end{align}
where $n$ is the number of nodes in the network, $c\in C$ denotes the node sequences sampled by the random walk, $b$ is the size of the window. The input data $R_{i}$ occupies the $i$-th row of the input matrix $R$. 
The node $v_{i+j}$ is the context node of the current node $v_i$ located in the generated random sequences in the window $b$. The node representation $y_{i}^{(K)}$ is the output of the sample input $R_i$ through the $K$ layer encoder. 
The matrix $H^{\prime}$ is the transition matrix between the representation output layer of the autoencoder and the output layer of the skip-gram, and $h_{i}^{\prime}$ is in Row $i$ of the transition matrix $H^{\prime}$.

Then, we obtain the following Equation (\ref{E4})
\begin{align}
\begin{split}
\label{E4}
L_{sg}^{NS}=-\sum_{i=1}^{n}\sum_{c\in C}\sum_{-b\le j\le b,j\ne 0}\{\log \sigma(h_{(i+j)}^{\prime }y_{i}^{(K)^ \mathrm{ T }}) +\\ \sum_{s=1}^{|neg|}\mathbb{E}_{v_n\sim P_n(v)} [\log \sigma(-h_{s}^{\prime }y_{i}^{(K)^ \mathrm{ T }})]\} ,
\end{split}
\end{align}
where $\sigma(x)=\frac{1}{1+\exp (-x)}$ is the sigmoid function, $|neg|$ denotes the number of the sampled negative samples. The sampling distribution $P_n(v)\propto d_v^{3/4}$ is set as suggested in \cite{Mikolov2013} and $d_v$ represents the degree of node $v_n$.

Minimizing Equation (\ref{E4}), 
we can get the result that if the two nodes co-occur, they have similar embedding vectors.

Algorithm \ref{alg:Framwork} describes the learning process of the entire model and all model parameters are denoted as $\Theta$.

\begin{algorithm}[htb]
	\caption{Framework of DANRL-ANE Model}
	\label{alg:Framwork}
	\begin{algorithmic}[1]
		\Require An attributed information network $G=(V,E,A,X)$, preprocessing hyperparameters $\eta$ and $\psi$, hadamard product operation parameter $\chi$, walks per node $r$, walk length $l$, window size $b$, return $p$, in-out $q$, negative samples $|neg|$, trade-off parameters $\alpha$ and $\beta$, regularizer coefficient $\gamma$, embedding dimension $d$
		\Ensure
		node vector representations $Y\in \mathbb{R}^{n \times d}$
		\State Use cosine similarity measurement method on attribute matrix to achieve attribute similarity matrix $X^S$
		\State Obtain the reconstructed adjacent matrix $R$ by linearly combining the adjacency matrix $A$ with attribute similarity matrix $X^S$ by $\eta$ and $\psi$
		\State Adopt random walk procedure of node2vec model with $p$ and $q$  both set as 1, and 
		start $r$ times of random walks with length $l$ at each node
		\State Random initialize all parameters $\Theta$
		\While {not converged}
		\State Sample a mini-batch of nodes with its context
		\State Compute the gradient of $\triangledown L_{FoP}$ based on Equation (\ref{E7})
		\State Update first-order proximity preserved module parameters
		\State Compute the gradient of $\triangledown L_{ae}^M$ based on Equation (\ref{E3}) and the gradient of $\triangledown L_{reg}$ based on Equation (\ref{E9})
		\State Update autoencoder module parameters
		\State Compute the gradient of $\triangledown L_{sg}^{NS}$ based on Equation (\ref{E4})
		\State Update skip-gram module parameters
		\EndWhile
		\State Obtain representations $Y=Y^{(K)}$ based on Equation (\ref{E20})
	\end{algorithmic}
\end{algorithm}

\begin{table}[h]
	\centering
	\begin{center}
		\caption{Dataset statistics}
		\begin{tabular}{|c|c|c|c|c|}
			\hline
			Datasets & $\#$ Nodes & $\#$ Edges & $\#$ Attributes & $\#$ Labels \\
			\hline
			Citeseer & 3,312 & 4,714 & 3,703 & 6 \\
			\hline
			PubMed & 19,717 & 44,338 & 500 & 3 \\
			\hline
			Cora & 2,708 & 5,429 & 1,433 & 7 \\
			\hline
			Facebook & 4,039 & 88,234 & 1,283 & - \\
			\hline
			Flickr & 7,575 & 239,738 & 12,047 & 9 \\
			\hline
		\end{tabular}
		\label{T1}
	\end{center}
\end{table}

\begin{table*}[h]
	\centering
	\begin{center}
		\caption{Basic network topology properties for datasets}
		\begin{tabular}{|c|c|c|c|c|}
			\hline
			Properties &  Density & Average degree & Average Clustering coefficient & Average distance \\
			\hline
			Citeseer & 0.0009 & 2.81 & 0.14 & unconnected\\
			\hline
			PubMed & 0.0002 & 4.50 & 0.06 & 6.34 \\
			\hline
			Cora & 0.0015 & 3.90 & 0.24 & unconnected \\
			\hline
			Facebook & 0.0108 & 43.69 & 0.61 & 3.69 \\
			\hline
			Flickr & 0.0084 & 63.30 & 0.33 & 2.41 \\
			\hline
		\end{tabular}
		\label{T2}
	\end{center}
\end{table*}

\section{Experiments}
In the section, compared with the state-of-the-art models, we verify the superiority of our proposed DANRL-ANE model via two downstream tasks, namely, the link prediction and node classification on five datasets, including Citeseer, PubMed, Cora, Facebook and Flickr.

\subsection{Experimental Settings}

\subsubsection{\textbf{Datasets}}

In the experiments, we select five public and widely used datasets, which belong to two network types, \textit{i.e.}, the citation networks and social networks. The 
dataset statistics are summarized in Table \ref{T1}. To further illustrate the broad applicability of our model to various networks, we analyze the basic topological properties of 
employed datasets, including the density, average degree, average clustering coefficient and average distance of the networks, as shown in Table \ref{T2}.

$\bullet$ \textbf{Citation networks}: Citeseer \cite{Zhang2018}, PubMed \cite{Zhang2018} and Cora \cite{Liu2019}.

The node indicates the publication, and the edge indicates the citing or cited relation between publications. 
By using the bag-of-words model to deal with a publication, and stemming and removing the stop-words, a vocabulary of the remaining unique words is used as the node attributes. 

In \textit{Citeseer}, publications are classified into six classes, namely, Agents, AI, DB, IR, ML and HCI; in \textit{PubMed}, publications are classified into  three classes, namely, Diabetes Mellitus Experimental, Diabetes Mellitus Type 1 and Diabetes Mellitus Type 2; and in \textit{Cora}, publications are classified into seven classes, namely, Case Based, Genetic Algorithms, Neural Networks, Probabilistic Methods, Reinforcement Learning, Rule Learning and Theory. The group categories are regarded as the labels of nodes.

$\bullet$ \textbf{Social networks}: Facebook \cite{Zhang2018} and Flickr \cite{Huang2017L}.

(1) \textit{Facebook}: It is one of the most famous online social networks. In the dataset, the node denotes the user, and the edge represents the friendship relation between two users. Furthermore, the personal profile is treated as the attribute information used to describe the user. Note that there are no labels in the dataset, so we cannot employ Facebook for the node classification.

(2) \textit{Flickr}: It is an image hosting and sharing website. Similarly, the node and the edge represent the user and the following or followed relation between users, respectively. The users can specify a list of tags that reflect their interests, which are processed into the attributes. The photos are organized under the pre-specified categories, so the labels refer to the photo interest groups that the users join in.

\subsubsection{\textbf{Baselines}}
We compare our model, DANRL-ANE, with seven state-of-the-art models, which can be divided as the following groups:

$\bullet$ \textbf{Structure-only:} The set of baseline models aim to capture the 
structural information, including: (i) the skip-gram based models which focus on preserving different order proximity of the structure, such as \textbf{DeepWalk}, \textbf{node2vec}, \textbf{LINE} and \textbf{GraRep}, (ii) the autoencoder based model, such as \textbf{SDNE}. Particularly,

(1) \textit{DeepWalk:} The truncated random walk is employed to capture the high-order proximity.

(2) \textit{node2vec:} The biased random walk is designed to explore the high-order structural information. 

(3) \textit{LINE:} The objective is to preserve the first-order and second-order proximity. Specifically, LINE models the direct and indirect neighbor relationship between node pairs through joint probability and conditional probability, respectively.

(4) \textit{GraRep:} All the local $k(k\ge 3)$-step relational information between node pairs are considered and concatenated as the final representations of nodes.

(5) \textit{SDNE:} Laplacian Eigenmaps and the deep model autoencoder are employed to preserve the first-order and second-order proximity, respectively.


$\bullet$ \textbf{Structure \& Attribute:} The models preserve the structural 
and attribute information based on the deep learning model autoencoder, 
which can be further classified: (i) the consistent learning based model, such as \textbf{DANE}, (ii) the deep coupling framework based model, such as \textbf{ANRL}.

(1) \textit{DANE:} The joint probability and autoencoder are used to mine the corresponding first-order and high-order proximity from the network structure, and to capture the corresponding first-order proximity and attribute semantics from the node attributes. Then, the likelihood estimation is used to learn the consistent network embedding from the structure and the attributes.

(2) \textit{ANRL:} It is a deep coupling model. The neighbor enhancement autoencoder module encodes the attribute semantics, and captures the second-order proximity. The attribute-aware skip-gram module is designed to preserve the high-order proximity. 
Furthermore, a large number of experiments in \cite{Zhang2018} have proved that in the ANRL variants, the performance of ANRL-WAN is superior. Hence, in the paper, we choose the ANRL-WAN as the benchmark.

\begin{table}[h]
	\centering
	\begin{center}
		\caption{Detailed architecture information for datasets}
		\begin{tabular}{|c|c|c|c|c|}
			\hline
			Datasets & $\#$ Neurons in each layer  \\
			\hline
			Citeseer & 3312—1000—500—128—500—1000—3312  \\
			\hline
			PubMed & 19717—1000—500—128—500—1000—19717  \\
			\hline
			Cora & 2708—1000—500—128—500—1000—2708 \\ & (Link Prediction) \\
			& 2708—256—128—256—2708 (Node Classification) \\
			\hline
			Facebook & 4039—1000—500—128—500—1000—4039(Link Prediction)  \\
			\hline
			Flickr & 7575—256—128—256—7575(Link Prediction) \\
			& 7575—500—128—500—7575(Node Classification)  \\
			\hline
		\end{tabular}
		\label{T3}
	\end{center}
\end{table}

\subsubsection{\textbf{Parameter settings}}

For all baselines, we use the public source code provided by the original author, and tune the parameters to make each model achieve the optimal performance on the different datasets and experimental tasks. We set the final embedding dimension $d$ as $128$. For \textit{LINE}, we concatenate the representations of the first-order and second-order proximity as the final embeddings. We set the walks per node $r$ as $10$, walk length $l$ as $80$, window size $b$ as $10$, negative samples $|neg|$ as $10$. 
The hyperparameters $\eta$, $\psi$, $\chi$, $\alpha$, $\beta$ and $\gamma$ are tuned by using the grid search. 
In addition, after the performance comparison of trying the application of $Relu$, $LeakyRelu$, $softsign$, $tanh$ and $sigmoid$ in our model, we use the $tanh$ function, $tanhx=\frac{sinhx}{coshx}=\frac{e^x-e^{-x}}{e^x+e^{-x}}$, as the non-linear activation function of the autoencoder. 
Table \ref{T3} shows the number of layers and dimension of each layer in the autoencoder, and there is no other layer between the representation layer of the autoencoder and the output layer of the skip-gram.

\begin{table}[htbp]
	\centering
	\begin{threeparttable} 
		\begin{center}
			\caption{Link prediction results on Citeseer, Pubmed, Cora, Facebook and Flickr datasets}
			\begin{tabular}{c|c|c|c|c|c}
				\hline
				\hline
				Datasets & Citeseer & PubMed & Cora & Facebook & Flickr \\
				\hline
				Evaluation & AUC & AUC & AUC & AUC & AUC \\
				\hline
				\hline
				DeepWalk & 0.6020 & 0.7925 & 0.7209 & 0.9461 & 0.7247 \\
				\hline
				node2vec & 0.5485 & 0.7977 & 0.7244 & 0.9552 & 0.7341 \\
				\hline
				LINE & 0.5309 & 0.6213 & 0.6047 & 0.5073 & 0.5262 \\
				\hline
				GraRep & 0.6008 & 0.8123 & 0.7210 & 0.8697 & 0.8899 \\
				\hline
				SDNE & 0.6093 & 0.7562 & 0.6326 & 0.8689 & 0.9023 \\
				\hline
				\hline
				DANE & 0.6579 & 0.9140 & 0.7286 & 0.8780 & 0.6142  \\
				\hline
				ANRL-WAN & \textbf{0.9666} & 0.8035 & 0.9181 & 0.7698 & 0.7800 \\
				\hline
				\hline
				DANRL-ANE & 0.9573 & \textbf{0.9439} & \textbf{0.9279} & \textbf{0.9577} & \textbf{0.9371} \\
				\hline
				\hline
			\end{tabular}
			\begin{tablenotes}
				\item $\star$ We use bold to highlight the best performance.
			\end{tablenotes}
			\label{T5}
		\end{center}
	\end{threeparttable}
\end{table}

\subsection{Link Prediction}
Link prediction is a widely used task to evaluate the performance of network embedding\cite{Zhang2018N}, which refers to the task of predicting either missing interactions or links that may appear in future in an evolving network \cite{Goyal2018} \cite{Hamilton2017}.

As done in \cite{Grover2016}, we hold out 50\% existing edges as positive instances, and ensure that the remaining network is connected. Besides, we randomly generate the same number of nonexistent edges from the original network, which are as negative instances. The positive and negative instances, together, constitute the test set. Furthermore, we use the residual network to train the embedding models, which is to obtain the representation of each node. Then, these representations are utilized as the feature inputs to predict the unobserved edges. Inspired by \cite{Zhang2018}, in the link prediction experiment, we rank both the positive and negative instances according to the cosine similarity function, and employ the AUC\cite{Fawcett2006} index to judge the ranking quality. A higher score indicates that the network representation is more informative. 
The link prediction task is carried out on all datasets. The AUC value for each model on the Citeseer, PubMed, Cora, Facebook and Flickr dataset is summarized in Table \ref{T5}, and the best result is highlighted in bold. According to the observations, we give the following analysis.

$\bullet$ \textbf{Structure vs.\ Structure:} 
On most datasets, node2vec achieves better or similar performance than DeepWalk, suggesting that the exploration of more flexible neighborhood facilitates the learning of node representations with higher accuracy. Compared with the models that only preserve partial microscopic structural information, such as DeepWalk, node2vec and LINE, the superior performance exhibited by GraRep in most of the time suggests that preserving the first-order, second-order, and higher-order proximity is necessary for the link prediction task. Furthermore, we compare the deep model SDNE and shallow model LINE, which both aim at capturing the first-order and second-order proximity. The result shows that SDNE consistently achieves better performance than LINE, especially on social networks. A comparison of the performance gap on citation networks and social networks, respectively, reveals that the larger the network average degree is, the smaller the performance gap is, which suggests that the autoencoder can capture the second-order proximity that is beneficial to link prediction.

$\bullet$ \textbf{Structure vs.\ Structure \& Attribute:} 
We find that the models considering both the structure and attributes tend to perform better than those considering only the structure. Furthermore, our proposed DANRL-ANE model has always achieved better performance than GraRep, especially on citation networks, which is worth noting that the node attributes in the PubMed dataset are relatively sparse. The above phenomenon shows that incorporating the attributes into NRL in a reasonable way is beneficial to obtain the excellent link prediction results.

$\bullet$ \textbf{Structure \& Attribute vs.\ Structure \& Attribute:} 
The comparison among DANRL-ANE, DANE and ANRL-WAN shows that DANRL-ANE has better experimental results on almost all datasets, which further proves the importance of capturing the first-order, second-order, high-order proximity and the attribute semantics together. Meanwhile, it demonstrates that DANRL-ANE can learn the robust and efficient network representation. 

\subsection{Node Classification}
Node classification aims to predict the categories of nodes by any known information of the network, 
which is another common downstream task to evaluate the performance of network embedding\cite{Zhang2018N}\cite{Goyal2018} \cite{Hamilton2017}. 

In the experiment, we first learn the vector representation of each node through different models. Then, following the popular practices\cite{Zhang2018}, we randomly sample 30\% nodes from the labeled nodes as the training set, and treat the rest as the test set. Here, SVM is employed as the classifier. To measure the multilabel classification performance, we use Micro-F1 and Macro-F1 as the evaluation metrics. Notably, the above classification process is repeated 10 times and we report the average results. 
Because we don't have the ground truth of node labels in Facebook, the node classification task is only performed on four datasets, \textit{i.e.}, Citeseer, PubMed, Cora and Flickr. Furthermore, Table \ref{T4} shows the performance of each network embedding method on different datasets, in which the optimal result is strengthened in bold. We analyze the results as follows.

$\bullet$ \textbf{Structure vs.\ Structure:} 
Node2vec also shows similar or superior results to DeepWalk on different datasets, which is the same as the previous experiments. Compared with LINE, SDNE shows poor performance on citation networks, especially in the PubMed dataset. Analyzing the characteristics of these networks, we find that the average degree of a network is a key factor affecting the performance of SDNE, and the larger the network average degree is, the worse the performance of SDNE is. The observation indicates that modeling the directly connected relationship between two nodes with joint probability is beneficial to capture the accurate first-order proximity, which explains why the joint probability is used in our model. Unlike that in link prediction, GraRep, which considers all the microscopic structural information, has worse performance than DeepWalk, node2vec and LINE, on citation networks. The result reveals that simply concatenating different order information is not always suitable for any tasks, which emphasizes that a careful design is critical.

$\bullet$ \textbf{Structure vs.\ Structure \& Attribute:} 
Table \ref{T4} shows that the methods incorporating the node attributes into NRL have better experimental results than those only focusing on the structure, which demonstrates the integration of the structural information and attribute semantics is advantageous to learn informative node vectors. For PubMed, compared with DeepWalk, the performance of DANE is poor, which is probably caused by the sparse attributes. 

$\bullet$ \textbf{Structure \& Attribute vs.\ Structure \& Attribute:} Based on the above discussion, the optimal result of DANRL-ANE for PubMed shows that our method is not susceptible to the sparsity of either network structure or node attributes. 
Meanwhile, the superiority of the proposed DANRL-ANE model over ANRL-WAN and DANE on almost all datasets proves that our method could learn the robust and efficient network representation, and explains the necessity of preserving the first-order, second-order and high-order proximity, which is noteworthy that the Citeseer and Cora dataset are both disconnected networks.
In a word, in the node classification task, the proposed DANRL-ANE model is applicable to all kinds of networks, even on sparse networks or networks with isolated nodes, if we can obtain the sufficient attribute information. 

\begin{table*}[htbp]
	\centering
	\begin{threeparttable}
		\begin{center}
			\caption{Node classification results on Citeseer, Pubmed, Cora and Flickr datasets}
			\begin{tabular}{c|c|c|c|c|c}
				\hline
				\hline
				Datasets & Citeseer & PubMed & Cora &  Flickr \\
				\hline
				Evaluation & Micro-F1 Macro-F1 & Micro-F1 Macro-F1 & Micro-F1 Macro-F1  & Micro-F1 Macro-F1 \\
				\hline
				\hline
				DeepWalk & 0.5665 \quad 0.5212 & 0.8109 \quad 0.7978 & 0.7900 \quad 0.7782 & 0.4940 \quad 0.4835 \\
				\hline
				node2vec & 0.6002 \quad 0.5465 & 0.8104 \quad 0.7968 & 0.8058 \quad 0.7942 & 0.5155 \quad 0.5062 \\
				\hline
				LINE & 0.5605 \quad 0.5256 & 0.8049 \quad 0.7926 & 0.7884 \quad 0.7767 & 0.5613 \quad 0.5576 \\
				\hline
				GraRep & 0.4775 \quad 0.4352 & 0.7416 \quad 0.7248 & 0.7636 \quad 0.7496 & 0.5692 \quad 0.5606 \\
				\hline
				SDNE & 0.4161 \quad 0.3632 & 0.4258 \quad 0.2900 & 0.5813 \quad 0.5201 & 0.6043 \quad 0.5991 \\
				\hline
				\hline
				DANE & 0.6870 \quad 0.6433 & 0.8063 \quad 0.7940 & 0.8110 \quad 0.7944 & 0.7721 \quad 0.7701 \\
				\hline
				ANRL-WAN & 0.7246 \quad \textbf{0.6764} & 0.8595 \quad 0.8584 & 0.8161 \quad 0.8030 & 0.6701 \quad 0.6584 \\
				\hline
				\hline
				DANRL-ANE & \textbf{0.7248} \quad 0.6744 & \textbf{0.8745} \quad \textbf{0.8728}  & \textbf{0.8291} \quad \textbf{0.8166} & \textbf{0.9059} \quad \textbf{0.9046} \\
				\hline
				\hline
			\end{tabular}
			\begin{tablenotes}
				\item $\star$ We use bold to highlight the best performance.
			\end{tablenotes}
			\label{T4}
		\end{center}
	\end{threeparttable}
\end{table*}

\section{Conclusion} 
To integrate the microscopic structural and attribute information for learning the robust and effective node embeddings from various networks, we propose a deep coupling model DANRL-ANE, where three newly designed modules are used to preserve the first-order, second-order and high-order proximity from the structure, respectively. In particular, the node attributes are incorporated into the adjacency matrix based on the social homophily, as the input of our model, so that the structure and attribute information are explored simultaneously. The extensive experiments on the tasks of link prediction and node classification show that our DANRL-ANE model achieves the superior performance comparing with other representation learning models. The work demonstrates that integrating more sources of information in a principled manner is conducive to learning higher quality network representation. 

\ifCLASSOPTIONcaptionsoff
  \newpage
\fi



\bibliographystyle{IEEEtran}
\bibliography{reference}
%



%


\begin{IEEEbiography}[{\includegraphics[width=1in,height=1.35in,clip,keepaspectratio]{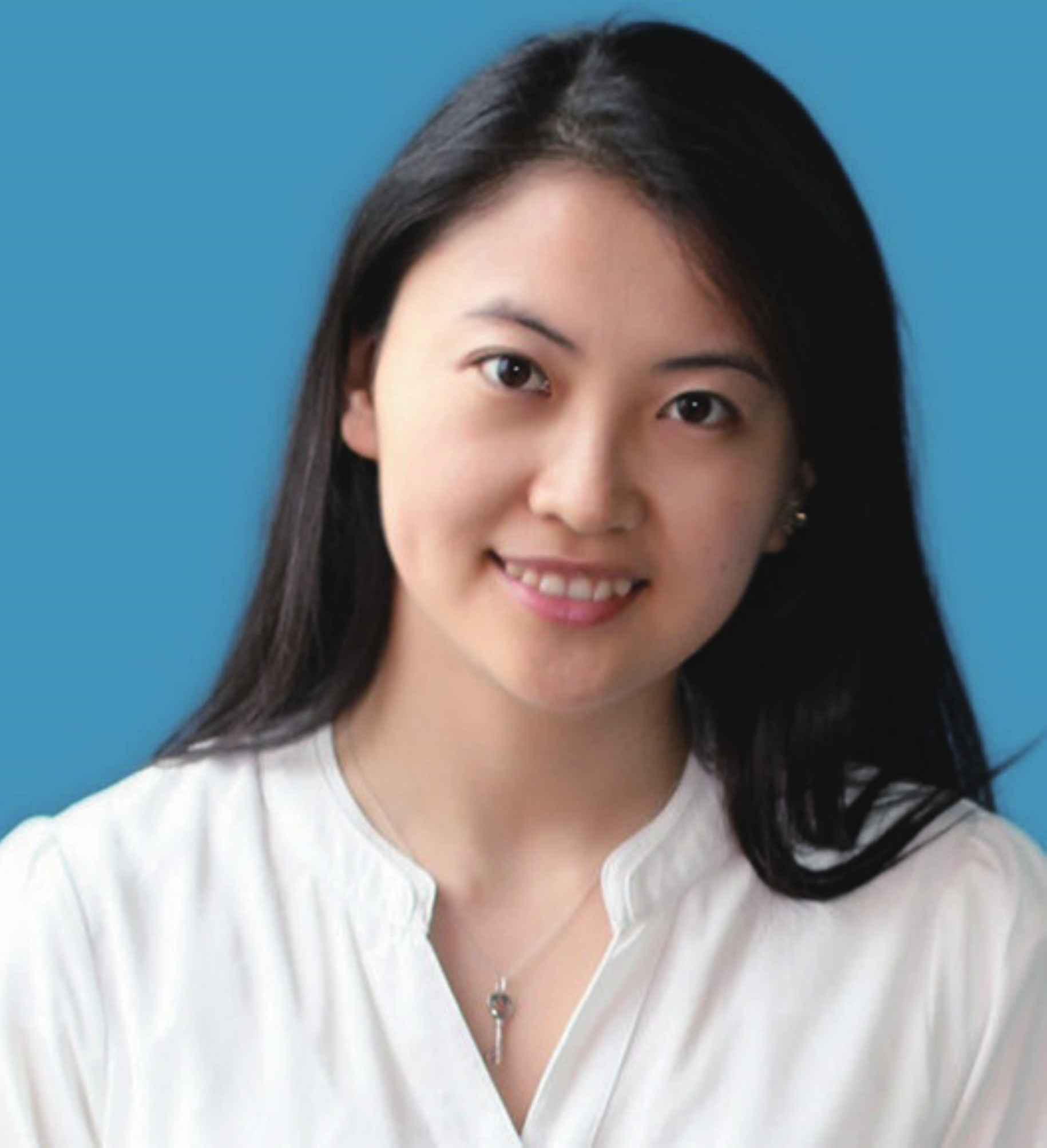}}]{Cong Li}
		(M'15) received the PhD degree in intelligent systems from Delft University of Technology (TUDelft), Delft, The Netherlands, in 2014. She is currently an associate professor in Electronic Engineering Department at Fudan University, where she involves in complex network theory and applications. Her research focuses on analysis and modeling of complex networks, including network properties, dynamic processes, network of networks etc. Her work in on these subjects include 2 (co-) authored research monographs, 1 book chapter and more than 30 international journal papers and conference papers.
\end{IEEEbiography}

\begin{IEEEbiography}[{\includegraphics[width=1in,height=1.2in,clip,keepaspectratio]{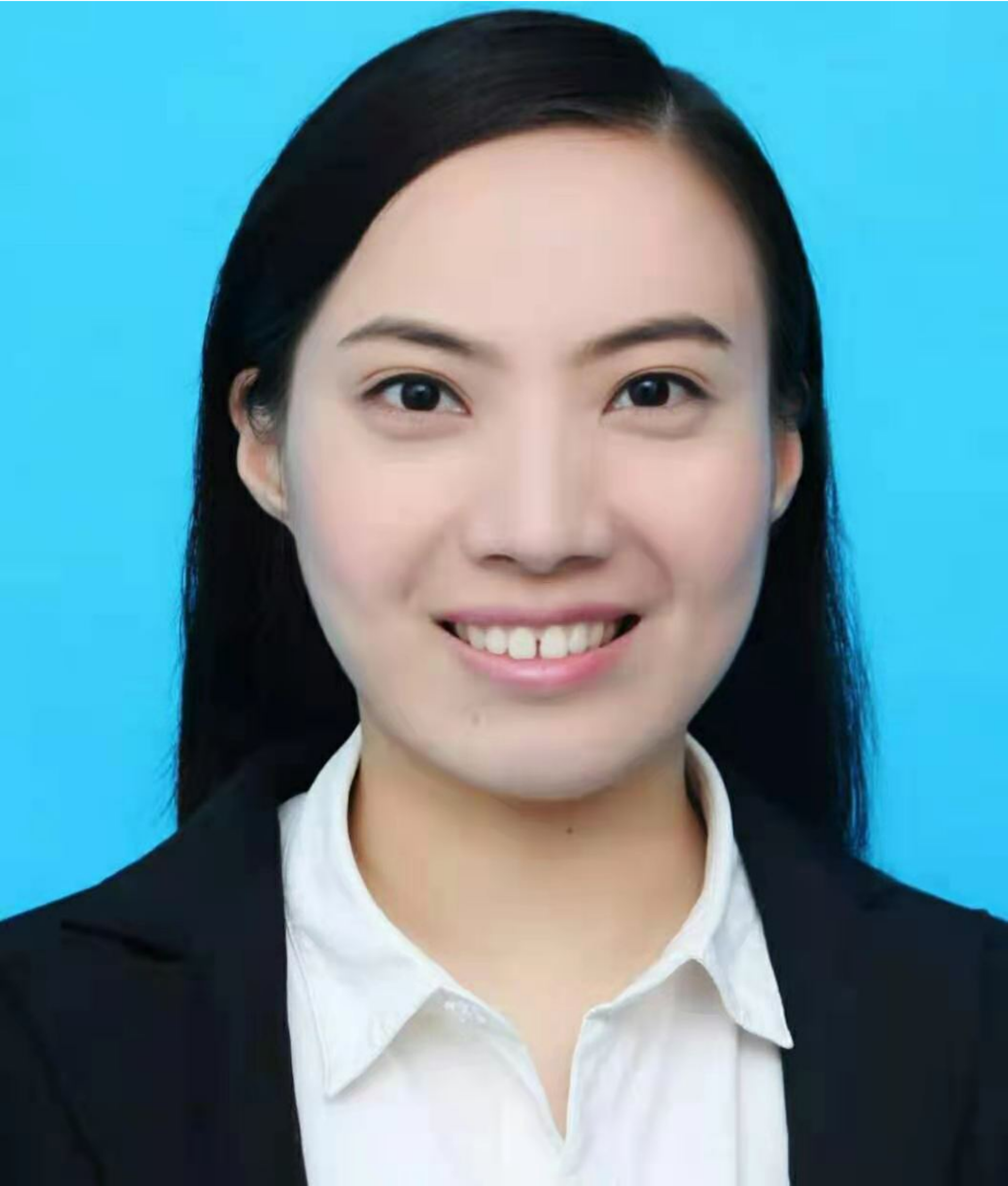}}]{Min Shi} received the BS degree in electronic information engineering from Nantong University, Jiangsu, China, in 2014. She is currently pursuing the M.S. degree with the Department of Information Science and Engineering, Fudan University, Shanghai, China. Her current research interests include data mining and network representation learning.
\end{IEEEbiography}

\begin{IEEEbiography}[{\includegraphics[width=1in,height=1.35in,clip,keepaspectratio]{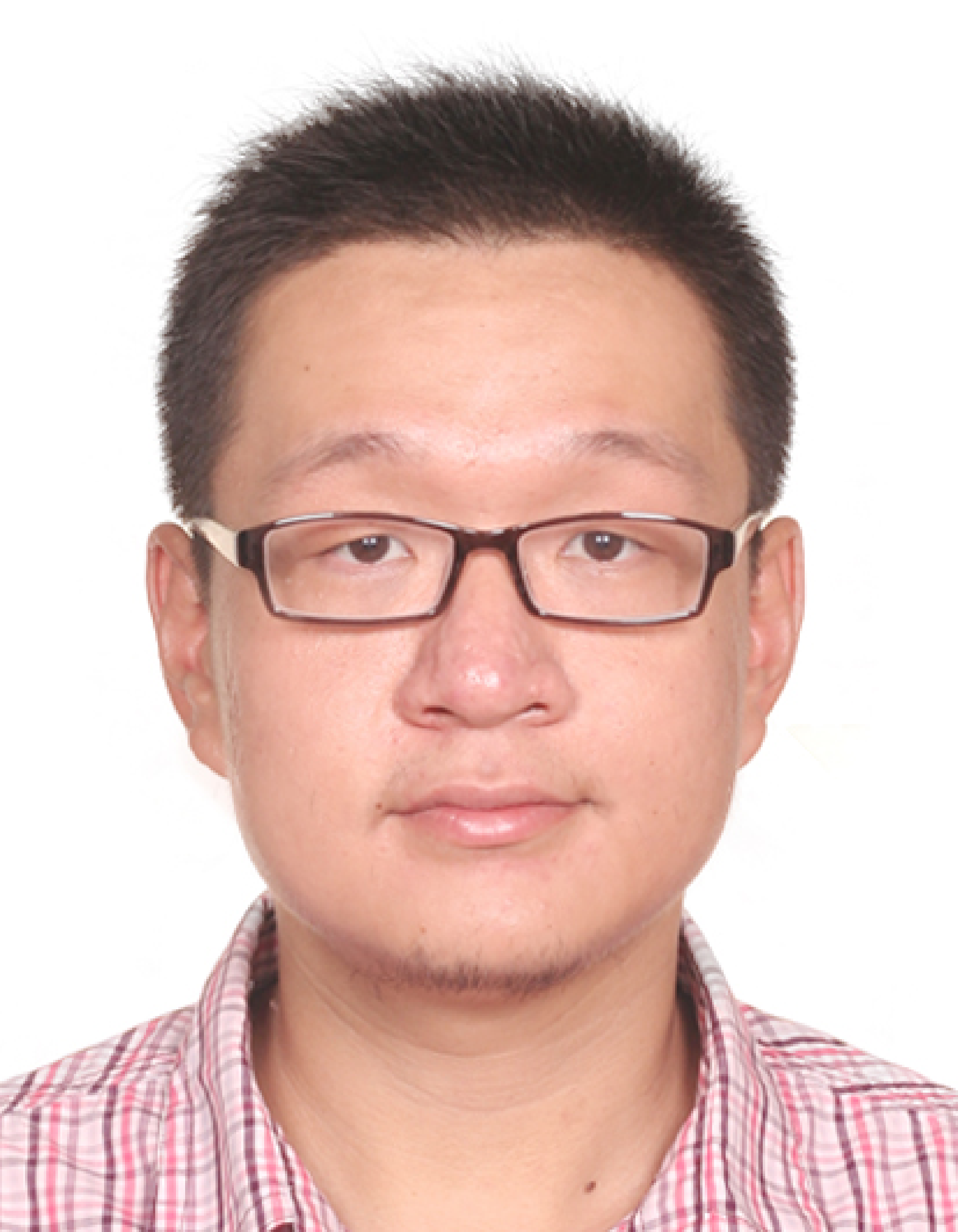}}]{Bo Qu} recieved the BS and MS degrees in information security and computer science from Shanghai Jiaotong University, China, in 2009 and 2012, respectively. He recieved the PhD degree in intelligent systems from Delft Univeristy of Technology, The Netherlands, in 2017. Before joining Peng Cheng Labortory as a research associate, he was with Tencent Technology as a researcher in applied research of network security. His research fouces on network security by network ananlysis, network representation learning, etc.
\end{IEEEbiography}

\begin{IEEEbiography}[{\includegraphics[width=1in,height=1.25in,clip,keepaspectratio]{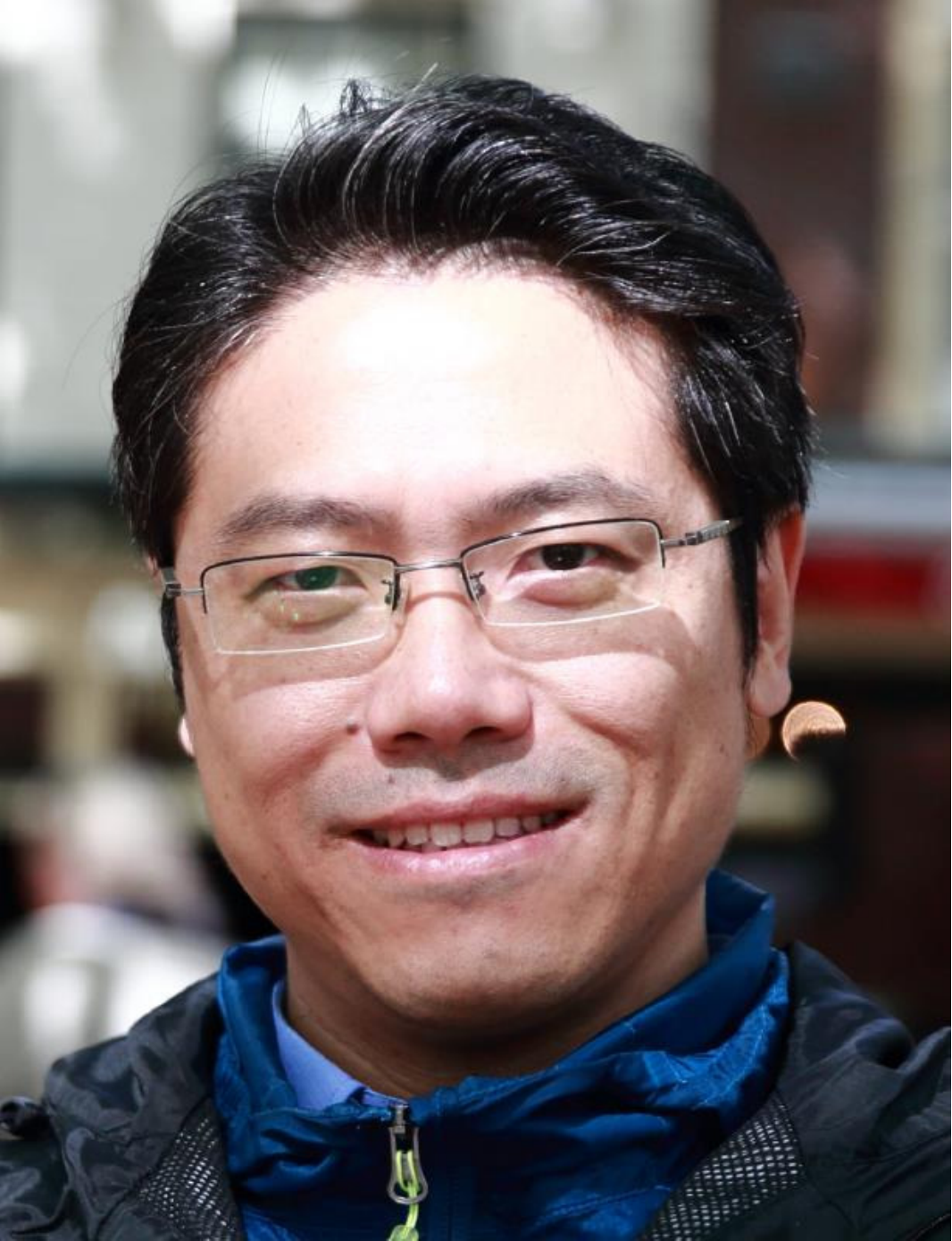}}]{Xiang Li}
	(M'05--SM'08) received the BS and PhD degrees in control theory and control engineering from Nankai University, China, in 1997 and 2002, respectively. Before joining Fudan University as a professor of the Electronic Engineering Department in 2008, he was with City University of Hong Kong, Int. University Bremen and Shanghai Jiao Tong University, as post-doc research fellow, Humboldt research fellow and an associate professor in 2002-2004, 2005-2006 and 2004-2007, respectively. He served as head of the Electronic Engineering Department at Fudan University in 2010-2015. Currently, he is a distinguished professor of Fudan University, and chairs the Adaptive Networks and Control (CAN) group and the Research Center of Smart Networks \& Systems, School of Information Science \& Engineering, Fudan University. He served as associate editor for the IEEE Transactions on Circuits and Systems-I: Regular Papers (2010-2015), and serves as associate editor for the IEEE Transactions on Network Science and Engineering, the Journal of Complex Networks and the IEEE Circuits and Systems Society Newsletter. His main research interests cover network science and systems control in both theory and applications. He has (co-)authored 5 research monographs, 7 book chapters, and more than 200 peer-refereed publications in journals and conferences. He received the IEEE Guillemin-Cauer Best Transactions Paper Award from the IEEE Circuits and Systems Society in 2005, Shanghai Natural Science Award (1st class) in 2008, Shanghai Science and Technology Young Talents Award in 2010, National Science Foundation for Distinguished Young Scholar of China in 2014, National Natural Science Award of China (2nd class) in 2015, Ten Thousand Talent Program of China in 2017, TCCT CHEN Han-Fu Award of Chinese Automation Association in 2019, among other awards and honors.
\end{IEEEbiography}







\end{document}